\documentclass{article}
\usepackage{ijcai21}

\usepackage{times}
\usepackage{soul}
\usepackage{url}
\usepackage[hidelinks]{hyperref}
\usepackage[utf8]{inputenc}
\usepackage[small]{caption}
\usepackage{graphicx}
\usepackage{amsmath}
\usepackage{amsthm}
\usepackage{amssymb}
\usepackage{booktabs}
\usepackage{algorithm}
\usepackage{algorithmic}
\usepackage{multirow}
\usepackage[symbol]{footmisc}

\usepackage{courier}
\usepackage{color,xcolor}
\usepackage{epsfig}
\usepackage{graphicx}

\usepackage{adjustbox}
\usepackage{array}
\usepackage{booktabs}
\usepackage{colortbl}
\usepackage{float,wrapfig}
\usepackage{hhline}
\usepackage{multirow}
\usepackage{caption}
\usepackage{subfigure}

\usepackage{amsmath,amsfonts,amsthm,amssymb}
\usepackage{bm}
\usepackage{nicefrac}
\usepackage{microtype}

\usepackage{changepage}
\usepackage{extramarks}
\usepackage{fancyhdr}
\usepackage{lastpage}
\usepackage{setspace}
\usepackage{soul}
\usepackage{xspace}

\usepackage{enumerate}
\usepackage[draft]{todonotes} 

\usepackage{titlesec}

\usepackage{comment}
\usepackage{tikz}
\usetikzlibrary{bayesnet}
\definecolor{MyDarkBlue}{rgb}{0,0.08,0.45}

\title{Context-Aware Image Inpainting with Learned Semantic Priors}

\author{
Wendong Zhang$^1$\and
Junwei Zhu$^2$\and
Ying Tai$^2$\and
Yunbo Wang$^1$\footnote{Corresponding Authors}\and\\
Wenqing Chu$^2$\and
Bingbing Ni$^1$\footnotemark[1]\and
Chengjie Wang$^2$\And
Xiaokang Yang$^1$\\
\affiliations
$^1$MoE Key Lab of Artificial Intelligence, AI
Institute, Shanghai Jiao Tong University\\
$^2$Youtu Lab, Tencent\\
\emails
\{diergent, yunbow, nibingbing\}@sjtu.edu.cn
}

\begin{document}

\maketitle

\begin{abstract}
  Recent advances in image inpainting have shown impressive results for generating plausible visual details on rather simple backgrounds. However, for complex scenes, it is still challenging to restore reasonable contents as the contextual information within the missing regions tends to be ambiguous. To tackle this problem, we introduce pretext tasks that are semantically meaningful to estimating the missing contents. In particular, we perform knowledge distillation on pretext models and adapt the features to image inpainting. The learned semantic priors ought to be partially invariant between the high-level pretext task and low-level image inpainting, which not only help to understand the global context but also provide structural guidance for the restoration of local textures. Based on the semantic priors, we further propose a context-aware image inpainting model, which adaptively integrates global semantics and local features in a unified image generator. The semantic learner and the image generator are trained in an end-to-end manner. We name the model SPL to highlight its ability to \emph{learn and leverage semantic priors}. It achieves the state of the art on Places2, CelebA, and Paris StreetView datasets\footnote[2]{Code available at \url{https://github.com/WendongZh/SPL}}. 
\end{abstract}

\section{Introduction}

Image inpainting aims at generating visually plausible image structures and local details for missing regions. This task has been researched for a long time and can be applied to many applications such as image restoration, object removal, and image manipulation~\cite{criminisi2004region}.
In summary, the main challenge for restoring realistic and reasonable contents lies in how to resolve the context ambiguities in missing regions. Although some approaches exploit stochastic models to learn the multi-modal distributions of the missing contents~\cite{zheng2019pluralistic,zhao2020uctgan}, in this paper, we focus on learning deterministic models that can leverage global context cues to reduce the context ambiguities.

\begin{figure}[h]
\begin{center}
\includegraphics[width=1.0\linewidth]{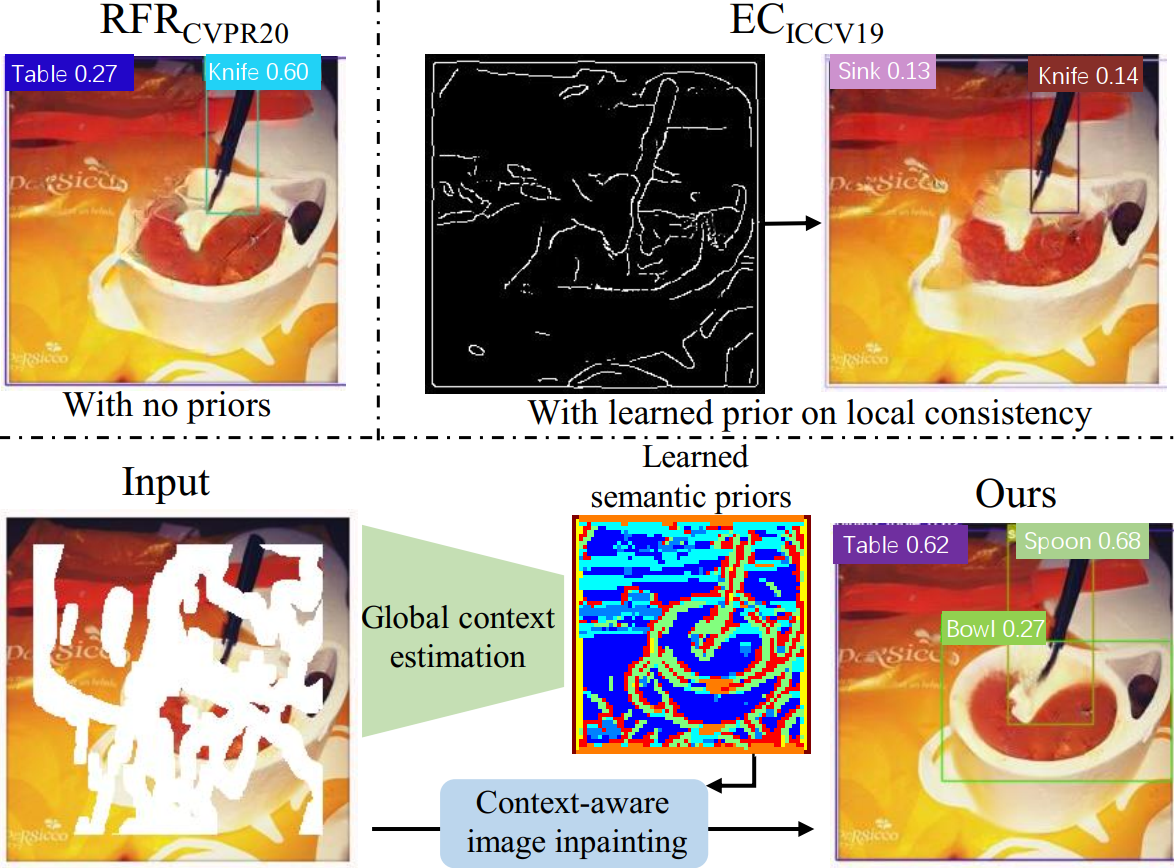}
\end{center}
\vspace{-4mm}
\caption{Illustration of SPL. Unlike existing methods (\textit{e.g.}, RFR~\protect\cite{li2020recurrent} and EC~\protect\cite{nazeri2019edgeconnect}) that mainly focus on local textures and are hard to restore reasonable contents especially when there exist context ambiguities, our model tackles this problem from a more semantic view. The detection results on the generated images show the benefits of learning semantic priors.}
\vspace{-4mm}
\label{motivation}
\end{figure}

In the realm of deterministic image inpainting, previous methods generally focus more on local texture consistency. The early work in this field usually uses the local texture similarities to perform pixel interpolations, such as texture diffusion or patch-based methods~\cite{barnes2009patchmatch,DBLP:conf/iccv/LevinZW03}. However, these methods may fail to restore non-repetitive structures.
Recently, many learning-based approaches are proposed in form of deep neural networks and directly reformulate the image inpainting as a conditional image generation problem~\cite{yu2020region,li2020recurrent,xie2019image,nazeri2019edgeconnect,DBLP:conf/eccv/LiuJSHY20}. 
Among these methods, some approaches further use particular structural supervisions to enhance the structure completion in final results, such as edge image~\cite{nazeri2019edgeconnect} and smooth image~\cite{DBLP:conf/eccv/LiuJSHY20}. 
Although these learning-based methods produce more realistic local details compared with early work, restoring reasonable image contents for more complex scenes is still hard as shown in Figure~\ref{motivation}. The main reason lies in that these methods fail to understand the global context but mainly focus on local texture consistency, which results in the distorted global structure.

In contrast to the above literature, we suggest that the high-level semantics can be considered as common knowledge between specific pretext tasks and image inpainting and that neural networks pre-learned for the pretext tasks can provide strong global relational priors that can be used to infer low-level missing structures.
To properly model the global relations between the missing contents and the remainder, we take the high-level feature maps of a pretrained multi-label classification (detection) model as the semantic supervisions\footnote{Various pretext tasks can be used to provide semantic supervisions. In the rest of the paper, we mainly use multi-label classification as a typical pretext task without loss of generality.}. 
These feature maps contain discriminative pre-learned semantic information of multiple objects which can help the global context understanding. Besides, as the pretext tasks mainly focus on the extractions of high-level features, they improve the robustness of our model to local textures and structural defects. 
The learned contextual features are referred to as \emph{semantic priors}, and we visualize an example via the k-means algorithm in Figure~\ref{motivation}. 
Following this line, a novel context-aware image inpainting model is proposed which considers both local texture consistency and global semantic consistency. 
The whole pipeline of our method can be summarized into two phases. In the first phase, we exploit two different encoders to separately extract the low-level image features and learn high-level semantic priors. 
By performing knowledge distillation on networks from pretext tasks\footnote{We do not train or fine-tune these models directly on the image inpainting datasets, but only use existing models that were pretrained on the Open Images dataset~\cite{kuznetsova2018open}.}, we obtain the global context representations. 
In the second phase, the learned semantic priors are spatially injected into image features to help image completion.
By learning and leveraging semantic priors, our approach (SPL) not only strengthens the local consistency of the generated image but also facilitates the structural inference of the missing contents from the remainder.

Our contributions to image inpainting can be summarized as follows:
\begin{itemize}
\item We show that distilling high-level knowledge from specific pretext tasks can benefit the understanding of global contextual relations and is thus semantically meaningful to low-level image inpainting. It requires no additional human annotations on the image inpainting datasets.
\item We propose SPL, a new model for context-aware image inpainting, which adaptively incorporates the learned semantic priors and the local image features in a unified generative network. It is trained with the semantic prior learner in an end-to-end fashion. 
\item{SPL achieves state-of-the-art performance on three standard image inpainting datasets.}
\end{itemize}

\section{Related Work}

\paragraph{Rule-based image inpainting.} Early work for image inpainting usually designs particular rules on low-level image features to establish the pixel-level mappings. More specifically, exemplar-based approaches~\cite{criminisi2004region,barnes2009patchmatch} 
exploit the patch or pixel-level similarities to directly copy textures in visible regions into missing regions. On the other hand, fusion-based methods~\cite{ballester2001filling,DBLP:conf/iccv/LevinZW03} try to smoothly propagate proper image contents into missing regions based on neighboring visible pixels. Although these methods produce vivid textures for simple background completion, they usually generate non-realistic images containing repetitive patterns due to the lack of high-level semantics.

\begin{figure*}[t]
\centering
\includegraphics[width=1.0\linewidth]{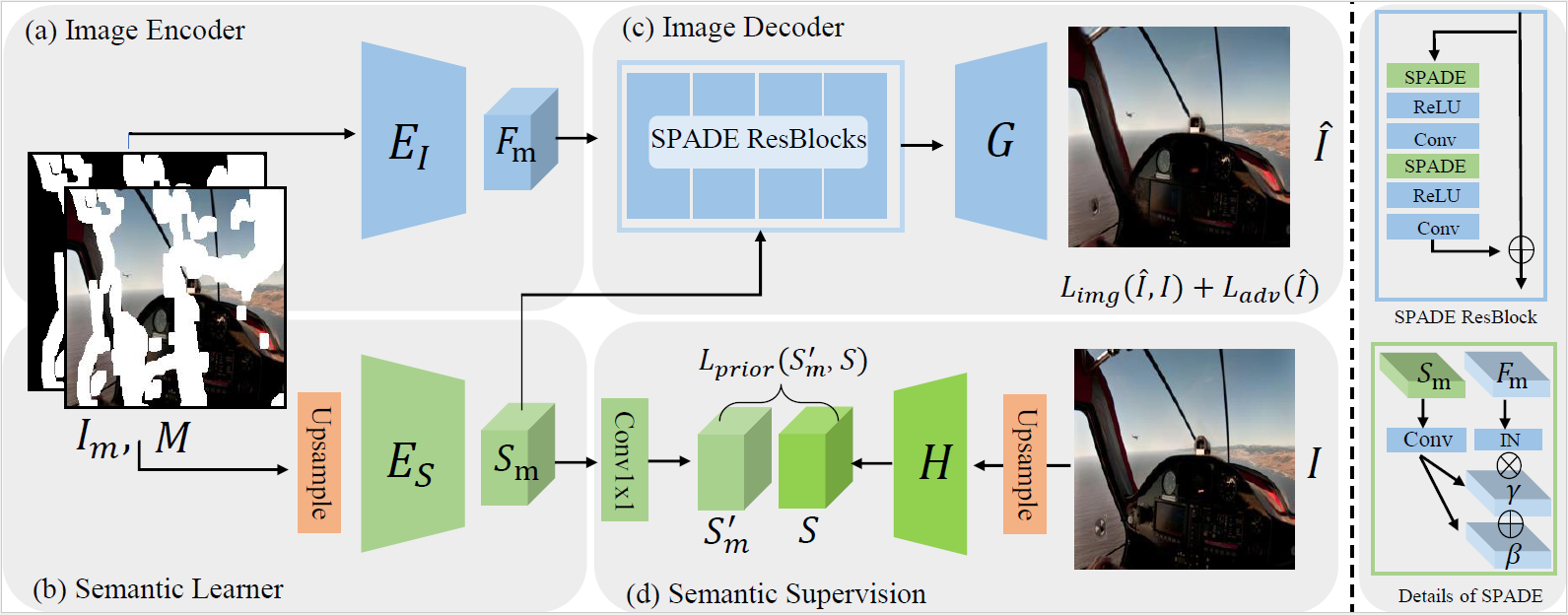}
\caption{Overview of SPL. In the encoding phase, it learns image features $\boldsymbol{F}_{m}$ and semantic priors $\boldsymbol{S}_{m}$ from input via the image encoder $E_{I}$ and the semantic learner $E_{S}$. In the generation phase, the semantic priors are spatially injected into the image features via spatially-adaptive normalization module (SPADE)~\protect\cite{park2019semantic} to help context-aware image inpainting. We exploit features $\boldsymbol{S}$ of a multi-label classification model ($H$)~\protect\cite{ben2020asymmetric} as the supervisions for leaning semantic priors, whose parameters are pretrained on the Open Image dataset and fixed during training. Modules (a-c) are trained in an end-to-end manner.  
}
\label{Overview}
\end{figure*}

\paragraph{Learning-based image inpainting.} Recent learning-based methods are mostly built upon generative adversarial networks \cite{goodfellow2014generative}, and formulate the image inpainting problem as conditional image generation. The work of context encoder \cite{pathak2016context} first utilizes an encoder-decoder architecture and proposes a channel-wise fully connected layer to estimate the missing contents. Subsequently, the global and local discriminators are introduced by \cite{iizuka2017globally} to further enhance the constraints of local consistency. After that, a two-stage coarse-to-fine generation framework is proposed by \cite{yu2018generative} with a context attention module. To exclude the influence of invalid pixels during the feature extraction process, the gated convolution layer~\cite{yu2019free} and the region normalization block~\cite{yu2020region} are proposed successively. Besides, the attention mechanism~\cite{xie2019image} and multi-scale learning framework~\cite{wang2019musical,li2020recurrent} are also introduced. Although these approaches use different network architectures for learning image features, they usually fail to restore reasonable structures in complex scenes due to the unresolved context ambiguities. 

\paragraph{Structure-enhanced image inpainting.} To resolve the structural distortion of the missing contents, some two-stage models~\cite{nazeri2019edgeconnect,ren2019structureflow} first generate edges or smooth images and then use them to enhance the structural details. 
Other approaches~\cite{DBLP:conf/eccv/LiuJSHY20,DBLP:conf/eccv/LiaoXWLS20} also use edge-preserving images or semantic maps as extra supervisions and require dense human annotations on the image inpainting dataset. 
In comparison, our model is trained on the image inpainting dataset in a fully unsupervised manner, and only exploits pre-learned features from the pretext tasks, where data annotations are more accessible.
Another difference is that, in our model, the semantic priors are learned and spatially injected into the image features in an end-to-end manner, which provides a more adaptive way for the restoration of the global and local structures.  

\section{Approach}
The overall architecture of the proposed approach is shown in Figure~\ref{Overview}. 
We name it SPL to highlight its ability to learn and leverage semantic priors. It contains three main components: an image encoder, a semantic priors learner, and an image decoder. In summary, the whole pipeline can be divided into two phases. In the encoding phase, we learn image features and semantic priors from the remainder of corrupted images. In the generation phase, the learned semantic priors are spatially incorporated into the image features to help the restoration of the global and local structural contents.

\subsection{Image Encoding}
Given corrupted image $\boldsymbol{I}_{m}$ and corresponding mask $\boldsymbol{M}\!\in\! \{0, 1\}$ where $1$ represents the locations for invalid pixels and $0$ represents those for valid pixels, we first utilize two different encoders $E_{I}$ and $E_{S}$ to learn image feature $\boldsymbol{F}_{m}$ and semantic priors $\boldsymbol{S}_{m}$. As these two features contain different information, we exploit different strategies to design the architectures of these two encoders, and they do not share any parameters.

For image encoder $E_{I}$ shown in Figure~\ref{Overview}(a), we aim to extract rich low-level features from visible contents to form the conditional input. These features play important roles in restoring vivid local textures. Therefore we exploit two down-sample layers to embed corrupted input into image feature:
\begin{align}
    \boldsymbol{F}_{m} = E_{I}(\boldsymbol{I}_{m}, \boldsymbol{M}),
\end{align}
where $\boldsymbol{I}_{m}\!\in\!\mathbb{R}^{3\times H\times W}$, $\boldsymbol{M}\!\in\! \mathbb{R}^{1 \times H \times W}$, $\boldsymbol{F}_{m}\!\in\! \mathbb{R}^{c \times \frac{H}{4} \times \frac{W}{4}}$. The image feature $\boldsymbol{F}_{m}$ will be further fed to the generation phase to restore plausible visual details.

\subsection{Semantic Priors Learning} 
For the semantic learner $E_{S}$ shown in Figure~\ref{Overview}(b), we aim to learn complete semantic priors for corrupted visual elements with supervisions from pretrained deep neural networks. To understand the global context which usually contains multiple different objects, as shown in Figure~\ref{Overview}(d), we exploit the multi-label classification model $H$ pretrained on the Open Image dataset~\cite{kuznetsova2018open} with asymmetric loss (ASL)~\cite{ben2020asymmetric} to provide supervisions, and we do not fine-tune this model on our image inpainting datasets. For this encoder, the texture details are less important and the encoder needs to pay more attention to understanding the global context in the remainder of the corrupted images. To this end, we make two modifications. 

First, as we aim to learn semantic information from higher layers of deep neural networks, we up-sample the input images to preserve more local structures. More specifically, we up-sample the full image $\boldsymbol{I}\!\in\! \mathbb{R}^{3 \times H \times W}$ into a large one $\boldsymbol{I}^{\prime}\!\in\! \mathbb{R}^{3 \times 2H \times 2W}$, and the corrupted image $\boldsymbol{I}_{m}$ and corresponding mask $\boldsymbol{M}$ are also up-sampled in the same manner to form the input $\boldsymbol{I}^{\prime}_{m}$ and $\boldsymbol{M}^{\prime}$ for learning semantic priors. After that, we utilize $\boldsymbol{I}^{\prime}$ as the input to the pretrained multi-label classification model $H$ to extract feature maps $\boldsymbol{S}\!\in\! \mathbb{R}^{d \times \frac{H}{4} \times \frac{W}{4}}$ as the supervisions of the semantic priors: 
\begin{align}
    \boldsymbol{S} = H(\boldsymbol{I}^{\prime}).
\end{align}

Second, to establish the mappings between visible contents and semantic priors, we exploit three down-sampling layers followed by five residual blocks to form the semantic learner $E_{S}$. The semantic learner $E_{S}$ utilizes enlarged corrupted image and mask as input to generate the semantic priors $\boldsymbol{S}_{m}$:
\begin{align}
    \boldsymbol{S}_{m} = E_{S}(\boldsymbol{I}^{\prime}_{m}, \boldsymbol{M}^{\prime}),
\end{align}
where $\boldsymbol{I}^{\prime}_{m}\!\in\! \mathbb{R}^{3 \times 2H \times 2W}$, $\boldsymbol{M}^{\prime}\!\in\! \mathbb{R}^{1 \times 2H \times 2W}$, $\boldsymbol{S}_{m}\!\in\! \mathbb{R}^{c \times \frac{H}{4} \times \frac{W}{4}}$. We then exploit a $1\!\times\!1$ convolutional layer to adapt $S_{m}$ to the representations from the pretext task: 
\begin{align}
    \boldsymbol{S}^{\prime}_{m} = Conv_{1\times 1}(\boldsymbol{S}_{m}),
    \label{tab:eq1x1}
\end{align}
where $\boldsymbol{S}^{\prime}_{m}\!\in\!\mathbb{R}^{d \times \frac{H}{4} \times \frac{W}{4}}$. Finally, an $\ell_1$ reconstruction loss with extra constraints on missing regions is proposed to supervise the semantic learner:
\begin{align}
    \mathcal{L}_{\text{prior}} = ||(\boldsymbol{S}- \boldsymbol{S}^{\prime}_{m})\odot (1 + \alpha \boldsymbol{M}_{s})||_{1},
    \label{tab:semantic supervision}
\end{align}
where $\odot$ represents Hadamard product operator, $\alpha$ represents the extra weight for missing regions and $\boldsymbol{M}_{s}$ represents resized mask which has the same spatial size as $\boldsymbol{S}$. With Eq.~\ref{tab:eq1x1} and Eq.~\ref{tab:semantic supervision}, we can distill the knowledge that are useful for image inpainting and filter out the task-agnostic components in $S$.

\subsection{Context-Aware Image Inpainting}

In the generation phase, we exploit the image decoder shown in Figure~\ref{Overview}(c) to tackle another key problem that how to utilize the learned semantic priors $\boldsymbol{S}_{m}$ to help the image restoration process. As the image feature $\boldsymbol{F}_{m}$ and learned semantic priors $\boldsymbol{S}_{m}$ focus on different aspects of image contents, directly concatenating these features for feature fusion will not only disturbs local textures in visible regions but also affects the learning processes of the corresponding encoders. 
To this end, we exploit $8$ ResBlocks with the \textit{spatially-adaptive normalization module} (SPADE)~\cite{park2019semantic} to spatially inject the semantic priors into the decoding phase. In this case, our model provides adaptive modifications on image features and decreases the unnecessary influences. Instead of utilizing images or semantic maps as conditional inputs, in our model, the affine transformation parameters in SPADE are learned from semantic priors and we provide the architecture of this module in Figure~\ref{Overview}. In more details, the SPADE modules in our model first normalize the input image feature $\boldsymbol{F}_{m}$ with non-parametric instance normalization. Then, two different sets of parameters are learned from semantic priors $\boldsymbol{S}_{m}$ to perform the spatially pixel-wise affine transformations on image feature $\boldsymbol{F}_{m}$: 
\begin{align}
    [\boldsymbol{\gamma}, \boldsymbol{\beta}] = \texttt{SPADE}(\boldsymbol{S}_{m}), \\
    \boldsymbol{F}^{\prime}_{m} = \boldsymbol{\gamma} \cdot \texttt{IN}(\boldsymbol{F}_{m}) + \boldsymbol{\beta},
\end{align}
where $\boldsymbol{\gamma}, \boldsymbol{\beta}, \boldsymbol{F}_{m}, \boldsymbol{F}^{\prime}_{m}\!\in\! \mathbb{R}^{c \times \frac{H}{4} \times \frac{W}{4}}$ and $\texttt{IN}$ represents non-parametric instance normalization. 
Finally, the output image is generated from the modified image feature $\boldsymbol{F}^{\prime}_{m}$ with generator $G$ which contains two upsampling layers:
\begin{align}
    \boldsymbol{\hat{I}} = G(\boldsymbol{F}^{\prime}_{m}).
\end{align}

\paragraph{Novelty of the use of SPADE.} 
Although the SPADE ResBlocks are the core modules of SPL, they are used for different purposes and in different ways from the original SPADE model.
Unlike those used for image-to-image translation, here they are jointly trained with two separate encoders and integrate the learned low-level features (instead of random variables) with the distilled high-level priors (instead of the raw layout data). 
The contribution of SPL is to show that extracting high-level knowledge from specific pretext tasks can be helpful to low-level image inpainting, which has not been thoroughly studied by previous work.

\subsection{Loss Function}

We exploit the reconstruction and adversarial losses to train the whole model of SPL. For the reconstruction loss, we use $\ell_1$ loss for the reconstructed image and focus more on the content of missing regions:
\begin{align}
    \mathcal{L}_{\text{img}} = ||(\boldsymbol{I} - \boldsymbol{\hat{I}}) \odot (\boldsymbol{1} + \delta \boldsymbol{M})||_{1},
\end{align}
where $\odot$ represents the Hadamard product operator and $\delta$ represents the weight for missing regions.
To generate more plausible local details, we also utilize the adversarial loss to train our model, along with a discriminator in an iterative training scheme:
\begin{align}
    \mathcal{L}_{\text{adv}} = -\mathbb{E}_{\hat{I}} [\log(1-D(\boldsymbol{\hat{I}}))],
\end{align}
\begin{align}
    \mathcal{L}^{D}_{\text{adv}} = \mathbb{E}_{I} [\log D(\boldsymbol{I})] +
    \mathbb{E}_{\boldsymbol{\hat{I}}} [\log(1-D(\boldsymbol{\hat{I}}))].
\end{align}

The full objective function of SPL can be written as
\begin{align}
    \mathcal{L} = \lambda_{1}\mathcal{L}_{\text{img}} + \lambda_{2}\mathcal{L}_{\text{adv}} + \lambda_{3}\mathcal{L}_{\text{prior}}.
\end{align}

We tune the hyper-parameters with grid search, and finally obtain $\lambda_{1}=10$, $\lambda_{2} = \lambda_{3} = 1$, $\alpha = 3$ and $\delta = 5$.

\section{Experiments}

\subsection{Experimental Setups} 
\paragraph{Datasets.} 
We evaluate our approach on three datasets, including Places2~\cite{zhou2017places}, CelebA~\cite{liu2015deep}, and Paris StreetView~\cite{doersch2015makes}. 
We use the standard training set of Places2 with over $4$ million images and evaluate all models on its validation set with $36{,}500$ images.
The CelebA human face dataset contains over $160{,}000$ training images and about $19{,}900$ testing images. 
The StreetView dataset contains $15{,}900$ training samples and $100$ testing samples. 
We use the mask set of \cite{liu2018image} with $12{,}000$ irregular masks that are pre-grouped into six intervals according to the size of the masks ($0\%$-$10\%$, $10\%$-$20\%$, $\ldots$, $50\%$-$60\%$). 
All images in Places2 and StreetView are resized to $256 \times 256$. As for CelebA, we first apply center cropping and then resize the images to $256 \times 256$, which is a common practice of previous work~\cite{li2020recurrent,yu2020region}.

\paragraph{Compared methods.} We compare our method (SPL) with state-of-the-art methods including the \textit{region normalization model} (RN)~\cite{yu2020region}, the \textit{recurrent feature reasoning model} (RFR)~\cite{li2020recurrent} and the \textit{mutual feature equalization mode} (MFE)~\cite{DBLP:conf/eccv/LiuJSHY20}. Among these methods, RN and RFR mainly focus on the consistency of local textures, and MFE exploits smooth images as extra supervisions to enhance structural completion. For methods that provide pretrained models, we directly evaluate them in our settings. Otherwise, we retrain the models based on the provided code.

\paragraph{Implementation and training details.}
To provide supervisions for semantic priors, we use multi-label classification as the pretext task, in which the model was pretrained on the Open Image dataset~\cite{kuznetsova2018open} with the asymmetric loss (ASL)~\cite{ben2020asymmetric}. We do not fine-tune the model on any image inpainting datasets. 
We adopt the patch-based discriminator from previous work \cite{yu2020region,nazeri2019edgeconnect}.
We apply the Adam solver with $\beta_1=0.0$ and $\beta_2=0.9$ for model optimization. The initial learning rate is $1e^{-4}$ for all experiments and decayed to $1e^{-5}$ at different epochs for different datasets. For Places2 and CelebA, we decay the learning rate at $30$ epochs and further fine-tune the model for another $10$ epochs. For Paris StreetView, we decay the learning rate at $50$ epochs and fine-tune the model for $20$ epochs. The batch size is set to $8$ for all datasets and we train our model with two V100 GPUs. 

\begin{figure*}[t]
\centering
\includegraphics[width=0.95\linewidth]{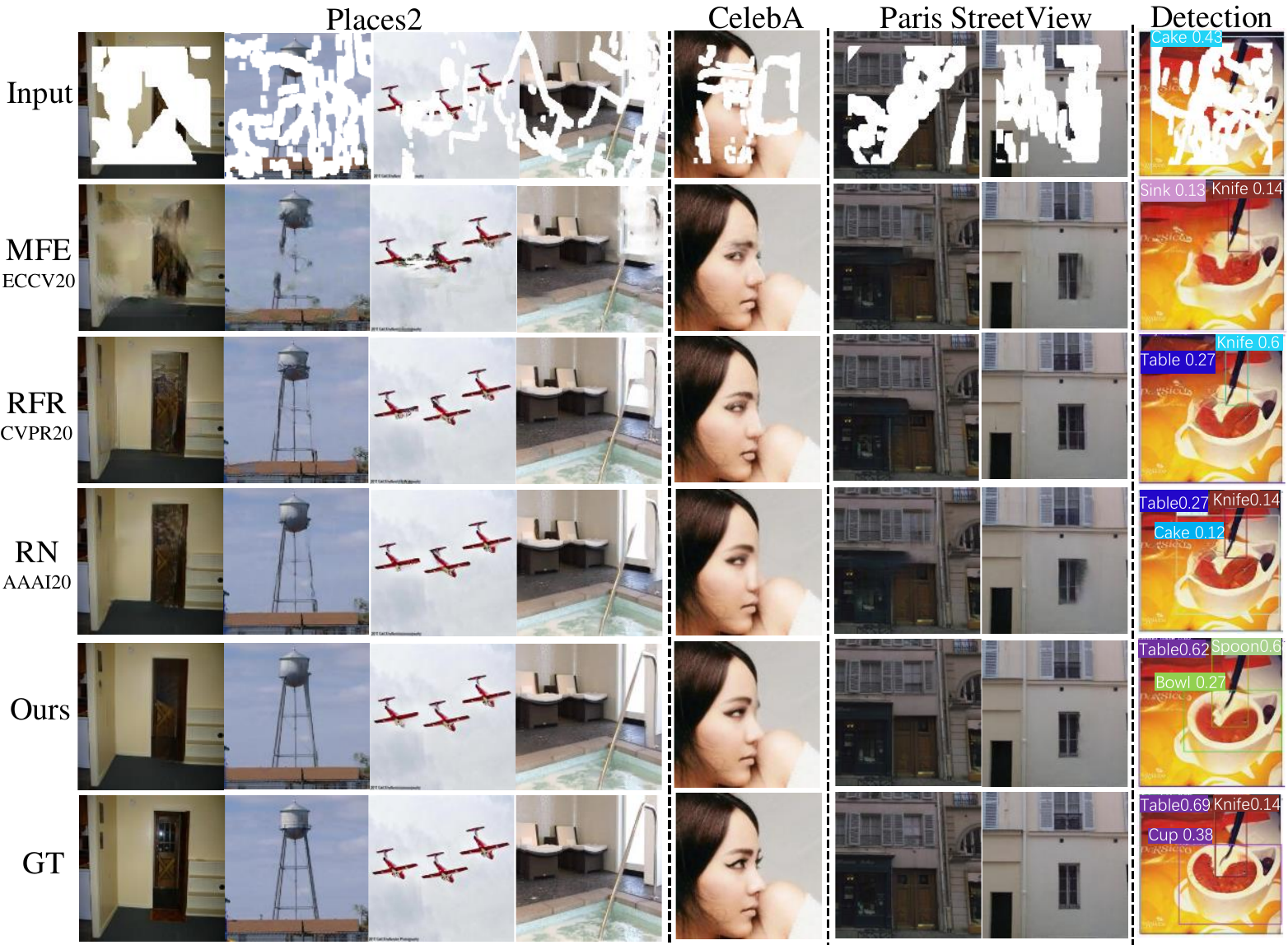}
\caption{Qualitative comparisons of our approach (SPL) with RN~\protect\cite{yu2020region}, RFR~\protect\cite{li2020recurrent}, and MFE~\protect\cite{DBLP:conf/eccv/LiuJSHY20}. Notably, as shown by the detection results on the generated images, SPL provides a better estimation of object-level contents and structures.}
\label{Qualititative}
\end{figure*}

\begin{table*}\footnotesize
	\centering
	\begin{tabular}{l|l|ccc|ccc|ccc}
		\hline
		\multicolumn{2}{c|}{Dataset} & \multicolumn{3}{c|}{Places2} & \multicolumn{3}{c|}{CelebA} & \multicolumn{3}{c}{Paris StreetView}\\
		\hline
		\multicolumn{2}{c|}{Mask Ratio} & 20\%-40\% & 40\%-60\% & All & 20\%-40\% & 40\%-60\% & All & 20\%-40\% & 40\%-60\% & All  \\
		\hline
		\hline
		\multirow{4}{*}{SSIM$^{\uparrow}$} 
		&MFE      & 0.820 & 0.655 & 0.809 & 0.916 & 0.806 & 0.900 & 0.872 & 0.707 & 0.834\\
		&RFR         & 0.858 & 0.706 & 0.842 & 0.931 & 0.837 & 0.917 & 0.893 & 0.763 & 0.863\\
		&RN        & 0.877 & 0.728 & 0.858 & 0.941 & 0.850 & 0.923 & 0.892 & 0.757 & 0.861\\
		&Ours      & \textbf{0.895} & \textbf{0.762} & \textbf{0.877} & \textbf{0.950} & \textbf{0.868} & \textbf{0.935} & \textbf{0.911} & \textbf{0.790} & \textbf{0.882}\\
		\hline
		\multirow{4}{*}{PSNR$^{\uparrow}$} 
		&MFE      & 23.07 & 19.22 & 24.24 & 29.01 & 23.78 & 30.14 & 27.22 & 22.07 & 27.07\\
		&RFR         & 24.52 & 20.49 & 25.72 & 30.30 & 25.02 & 31.39 & 28.32 & 23.71 & 28.36\\
		&RN          & 25.55 & 21.16 & 26.58 & 31.15 & 25.43 & 31.94 & 28.44 & 23.53 & 28.37\\
		&Ours        & \textbf{26.58} & \textbf{22.14} & \textbf{27.71} & \textbf{32.14} & \textbf{26.34} & \textbf{33.27} & \textbf{29.34} & \textbf{24.47} & \textbf{29.38}\\
		\hline
		\multirow{4}{*}{MAE $^{\downarrow}$} 
		&MFE       & 0.0260 & 0.0531 & 0.0287 & 0.0120 & 0.0289 & 0.0146 & 0.0159 & 0.0412 & 0.0226\\
		&RFR          & 0.0209 & 0.0435 & 0.0233 & 0.0099 & 0.0240 & 0.0122 & 0.0134 & 0.0324 & 0.0182\\
		&RN           & 0.0184 & 0.0403 & 0.0212 & 0.0091 & 0.0232 & 0.0119 & 0.0136 & 0.0337 & 0.0187\\
		&Ours         & \textbf{0.0161} & \textbf{0.0355} & \textbf{0.0187} & \textbf{0.0080} & \textbf{0.0205} & \textbf{0.0102} & \textbf{0.0115} & \textbf{0.0292} & \textbf{0.0161}\\
		\hline
	\end{tabular}
	\caption{Quantitative comparisons of the generated images by our approach (SPL) and the state of the art on three image inpainting datasets.
	}
	\label{tab:Quantitative_comparison in datases}
\end{table*}

\subsection{Qualitative Comparisons}
We provide qualitative results on three different datasets in Figure~\ref{Qualititative}. The Places2 dataset is the most challenging dataset as it contains different complex scenes and various objects. For this dataset, we can observe from the first four columns that results from our methods contain more complete global and local structures. %
On the contrary, results from other methods contain obvious distorted structures and suffer from blur effects. These results suggest that our method successfully reduces the context ambiguities in missing regions and obtains more complete visual concepts for foreground objects. 

For CelebA dataset, results are shown in the fifth column. In this case, all methods produce complete human faces as this dataset only contains various aligned human faces. However, as shown in the fifth column, previous approaches generate meaningless textures in local details. 
In other words, these methods fail to globally understand the context information and pay more attention to fill the missing regions with textures based on local distributions. Compared with them, our results contain more clear local details, which suggests that the learned semantic priors can also help local texture restoration. Similar observations can also be found in Paris dataset.

We further provide detection results of the output images in the last column in Figure~\ref{Qualititative}. From these results, we can observe that our result contains more complete objects with higher detection confidences than other methods.

\subsection{Quantitative Comparisons}
We further provide the quantitative comparison results in Table~\ref{tab:Quantitative_comparison in datases}. To quantitatively evaluate our performance, we select peak signal-to-noise ratio (PSNR), structural similarity (SSIM), and mean $\ell_{1}$ error (MAE) as our evaluation metrics which are commonly used in previous work. The evaluation process is as follows. We first randomly sample three different mask sets from the whole irregular mask dataset. These mask sets are assigned to different image datasets to form the mask-image pairs. For each image dataset, the mask-image mappings are held for different methods to obtain fair comparisons. The results of the whole testing sets are shown in columns with mask ratio being ``All'', and we gather those images with masks in corresponding mask ratios to calculate other results in Table~\ref{tab:Quantitative_comparison in datases}. We can see that our method obtain obvious improvements on all three metrics, which further demonstrates the effectiveness of our method.

\begin{figure}[t]
\vspace{2mm}
\begin{center}
\includegraphics[width=1.0\linewidth]{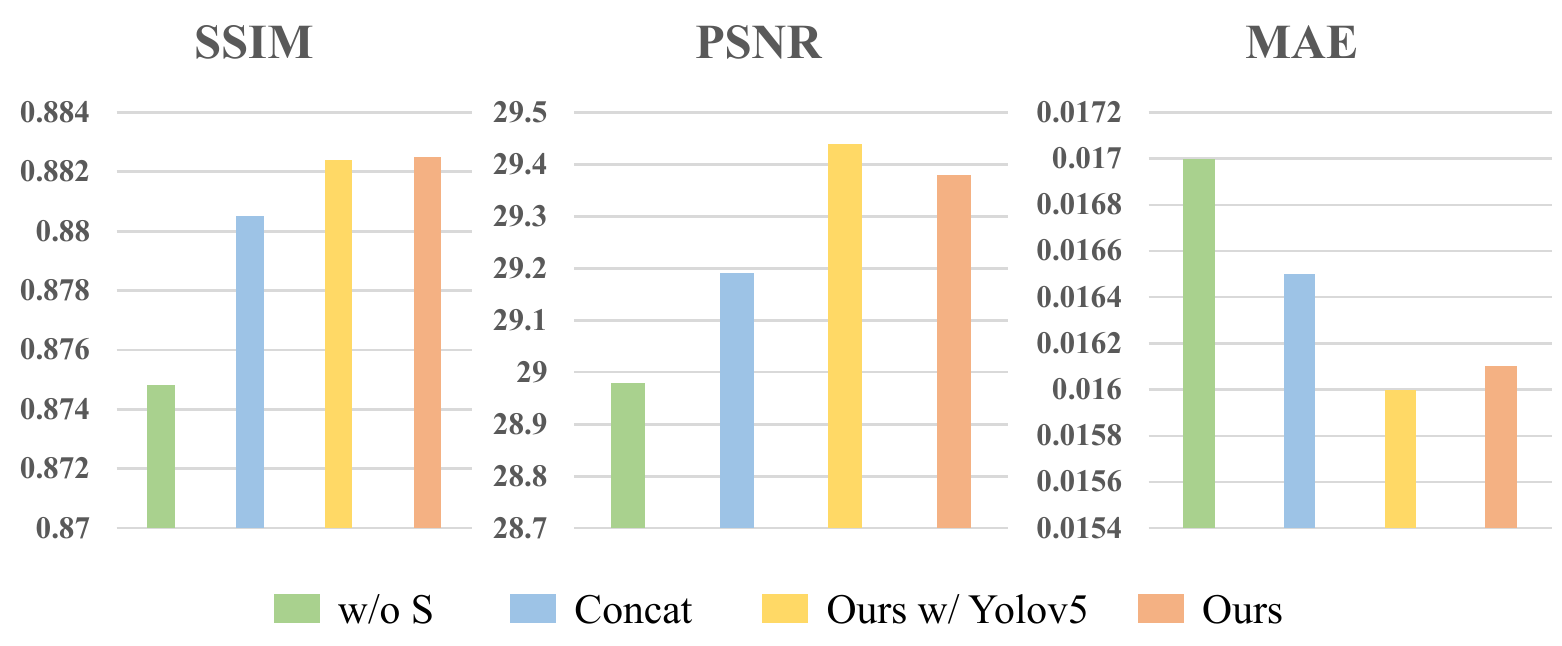}
\vspace{-5mm}
\end{center}
    \caption{Comparisons on the StreetView dataset of (1) different ways to integrate the learned semantic priors into the image decoder, and (2) using different semantic supervisions (vs. w/ YoloV5).} 
\label{img:ab}
\end{figure}

\begin{figure}[t]
\begin{center}
\includegraphics[width=1.0\linewidth]{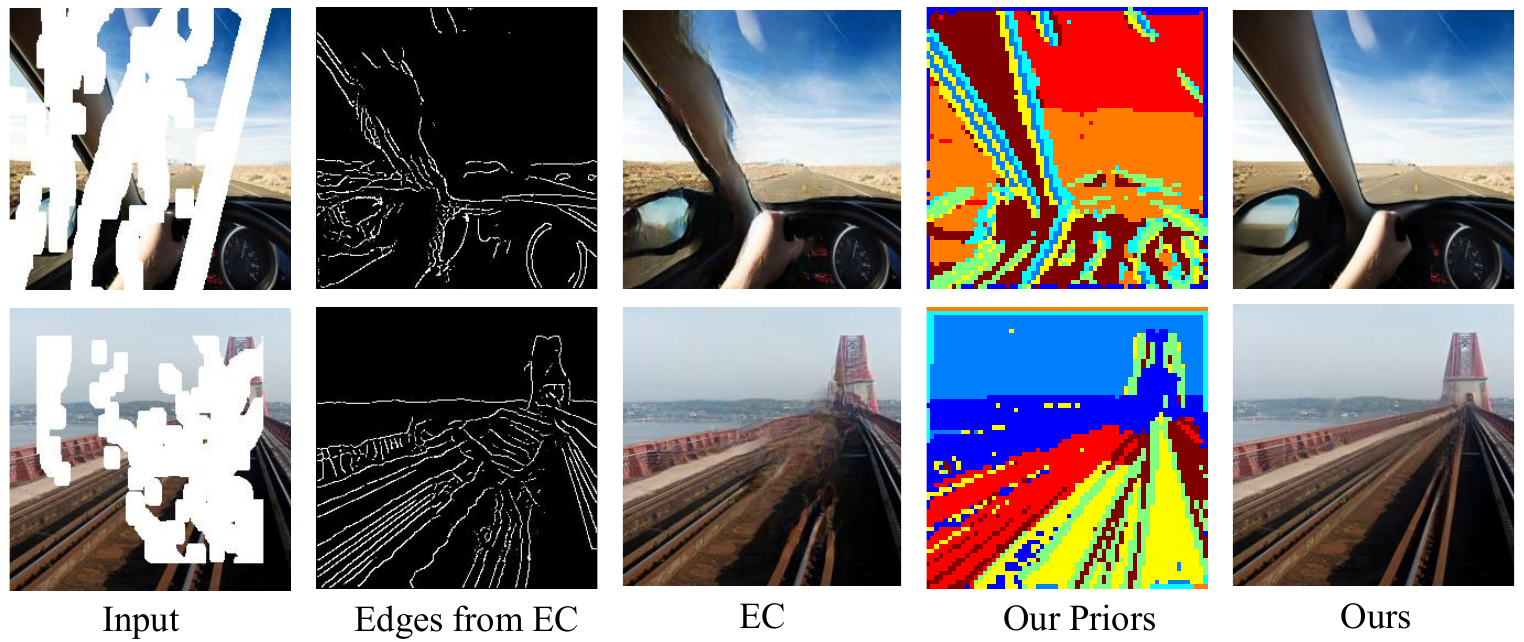}
\vspace{-5mm}
\end{center}
    \caption{Visualization of
    the semantic priors using the k-means algorithm, compared with \textit{edgeconnect} (EC)~\protect\cite{nazeri2019edgeconnect}. 
    }
\label{img:visualization}
\end{figure}

\subsection{Ablation Studies}
 
\paragraph{Alternatives of the semantic supervisions.} We then conduct two different experiments to evaluate the effects of the proposed semantic supervisions. First, we directly remove the semantic supervisions from our model, named as ``w/o S''. Then, we replace the pretrained multi-label classification network with a pretrained detection network YoloV5~\cite{web_reference} to provide semantic supervisions. In this case, we aim to explore whether other pretext models can also provide sufficient information for global context understanding. From Figure~\ref{img:ab}, we can obtain two observations: 1) The performance of our method will suffer from significant degeneration without explicit supervision from pretrained model. 2) Supervisions from the detection model can also provide helpful semantic information for global context understanding.

\paragraph{Alternatives of SPADE in SPL.} Instead of using the SPADE modules, in this experiment, we directly concatenate the learned semantic priors and the features from the image encoder to form the input of ResBlocks, which is commonly used by previous tow-stages generation frameworks~\cite{nazeri2019edgeconnect,ren2019structureflow}. Results in Figure~\ref{img:ab} show that the straightforward operation of concatenation leads to decreases in all three evaluation metrics.

\subsection{Visualization of Learned Semantic Priors} 
We further visualize the learned semantic priors and compare the generation results with the edge pre-generation method (EC)~\cite{nazeri2019edgeconnect} in Figure~\ref{img:visualization}. Specifically, we directly conduct the k-means algorithm on the learned semantic priors and use different colors to represent different clusters. From this figure, we can see that the EC model cannot restore reasonable structures for complex scenes. Besides, the distorted structures in edge images will further mislead the final results. Instead, the semantic priors learned by our method reduce the context ambiguities and provide explicit guidance for both global and local structure restoration. 

\section{Conclusion}
In this paper, we propose a new solution to image inpainting named SPL. Instead of focusing on the consistency of local textures, we show that the high-level features from deep neural networks, which were well-trained under specific pretext tasks (\textit{e.g.}, multi-label classification and object detection), contain rich semantic information and can benefit global context understanding. 
We transfer the high-level knowledge from the pretext models to low-level feature space and use the learned priors as the structural guidance of image inpainting.
Finally, we propose the context-aware image inpainting model that adaptively incorporates global semantics with local textures in a unified image decoding module. 
SPL significantly outperforms the state-of-the-art approaches on three datasets.

\section*{Acknowledgments} 
This work was supported by NSFC (U19B2035, U20B2072, 61976137). It was also supported by the Shanghai Municipal Science \& Technology Major Project (2021SHZDZX0102).


\bibliographystyle{named}
\bibliography{ijcai21}

\end{document}